\documentclass[twoside]{article}

\usepackage{graphicx}
\usepackage{xcolor}
\usepackage[accepted]{aistats2026}
\usepackage{comment}
\usepackage{booktabs}
\usepackage{siunitx}
\usepackage{makecell}
\usepackage{amsmath}  
\usepackage{amssymb}  
\usepackage{algorithm}
\usepackage{algorithmic}
\usepackage{natbib}
\usepackage{subcaption}
\usepackage{url} 
\DeclareMathOperator*{\argmin}{arg\,min}
\DeclareMathOperator*{\argmax}{arg\,max}
\begin{document}

%
\runningtitle{Black-Box Optimization From Small Offline Datasets via Meta Learning with Synthetic Tasks}

%

\twocolumn[

\aistatstitle{Black-Box Optimization From Small Offline Datasets \\ via Meta Learning with Synthetic Tasks}

\aistatsauthor{ Azza Fadhel \And The Hung Tran \And  Trong Nghia Hoang \And Jana Doppa}

\aistatsaddress{School of EECS, Washington State University, Pullman, WA, USA} ]

\begin{abstract}
We consider the problem of offline black-box optimization, where the goal is to discover optimal designs (e.g., molecules or materials) from past experimental data. A key challenge in this setting is data scarcity: in many scientific applications, only small or poor-quality datasets are available, which severely limits the effectiveness of existing algorithms.
Prior work has theoretically and empirically shown that performance of offline optimization algorithms depends on how well the surrogate model captures the optimization bias (i.e., ability to rank input designs correctly), which is challenging to accomplish with limited experimental data. This paper proposes {\em Surrogate Learning with Optimization Bias via Synthetic Task Generation} (\textsc{OptBias}), a meta-learning framework that directly tackles data scarcity. OptBias learns a reusable optimization bias by training on synthetic tasks generated from a Gaussian process, and then fine-tunes the surrogate model on the small data for the target task. 
Across diverse continuous and discrete offline optimization benchmarks, OptBias consistently outperforms state-of-the-art baselines in small data regimes. These results highlight OptBias as a robust and practical solution for offline optimization in realistic small data settings.
\end{abstract}

\vspace{-2ex}

\section{INTRODUCTION}
\label{sec:introduction}
Many science and engineering applications involve optimizing complex design spaces (e.g., materials, hardware, molecules) using data from expensive lab experiments or computational simulations. Some examples include discovering new materials~\citep{butler2018machine,deshwal2021bayesian,gantzler2023multi,fadhel2026nanoporous}, designing protein and drug molecules~\citep{vamathevan2019applications,Dara2021_PMC}, designing hardware~\citep{BOPS}, and optimizing parameters of additive manufacturing processes~\citep{Dara2021_PMC,fadhel2026discovery}. To avoid the prohibitively high cost of direct experimentation, a more affordable strategy is to explore designs {\em in silico} by learning a surrogate model from experimental data and subsequently optimizing it. Existing research in this area follows two directions: (1) online optimization, where new experiments are iteratively conducted to refine the surrogate model~\cite{Snoek12, SnoekICML15}; and (2) offline optimization, where the surrogate model is learned using only past experimental data~\citep{TrabuccoICML21, TrabuccoArXiv22}.


Among these approaches, offline optimization is an emerging paradigm that has become increasingly preferable in situations where the high overhead and limited throughput of physical experimentation often makes online optimization impractical~\citep{TrabuccoICML21}. The key challenge in offline optimization lies in the surrogate model’s unreliable predictions for out-of-distribution (OOD) inputs which are far from the offline data regime. As a result, iterative extrapolation of the learned surrogate during the search process (e.g., gradient-based search) suffers from error propagation as the search moves away from the offline dataset, leading to suboptimal designs.  

Existing research in offline optimization has largely focused on addressing this challenge using various conservative learning and optimization strategies which discourage the surrogate model from overestimating the output at unseen inputs from OOD regimes~\citep{trabucco2021conservative}. However, most of this work is empirical and often lack a principled foundation. The more recent works of~\cite{hoang2025learning, tan2024offline} have also pointed out that prior methods have inadvertently encoded an unnecessary value-matching bias in their surrogate modeling despite the simple intuition that shifting a function by a constant does not alter its optima. This observation suggests a key insight: rather than emphasizing value matching, optimization should emphasize gradient matching, which is more robust. The work of ~\cite{hoang2025learning} has in particular established that the optimization performance is bounded by the gradient gap between the surrogate and the unknown oracle black-box functions. This reveals and emphasizes on an important optimization bias which prefers surrogates that match the oracle's gradient field better when gradient based search is used for optimization.~\cite{hoang2025learning} also introduced \textsc{Match-Opt} which is a gradient matching algorithm: by observing empirical differences between offline data points,  interpolates the oracle's gradient field. This method is effective in the large offline data regime but struggles when the offline experimental data is limited, a common bottleneck in many real-world applications (e.g., drug design and materials discovery). For a conceptual illustration, we perform evaluation on the 4D Shekel synthetic function and observe that the gradient estimation error increases significantly as the the number of offline training examples reduce. This raises the main research question of this paper:

\begin{figure}[t]
  \centering
  \includegraphics[width=\columnwidth]{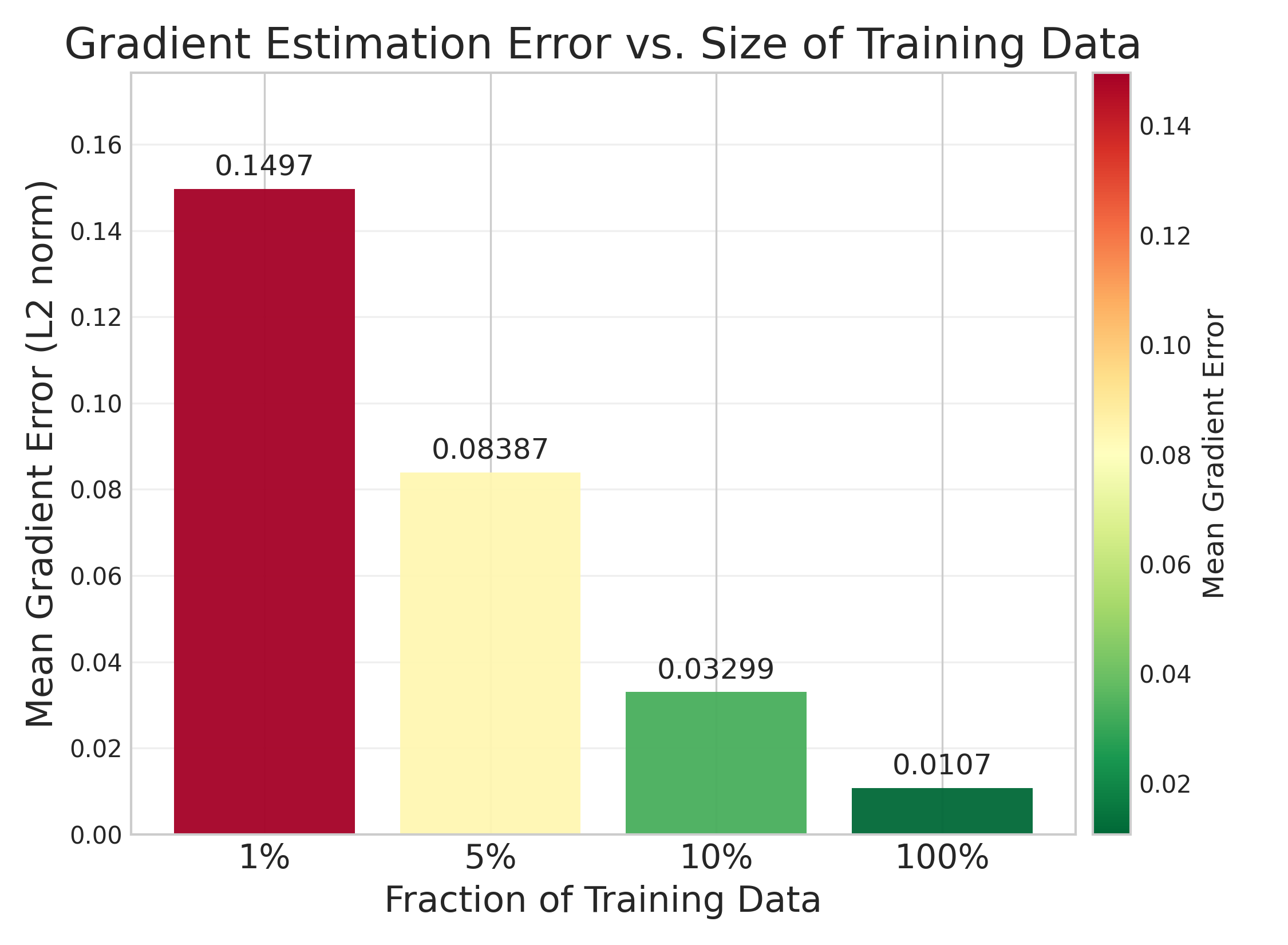}
\caption{Conceptual illustration of gradient estimation error vs.\ fraction of training data on the Shekel-4D synthetic task. The surrogate is trained on 8{,}000 samples and evaluated on 2{,}000 test points; bars are colored red→green for higher→lower error. The gradient estimation error increases significantly in the small training data regime relative to the full (100 \%) data.}
  \label{fig:grad-error-bar}
  \vspace{-3ex}
\end{figure}

{\em How to learn a surrogate to effectively capture optimization bias when offline data are limited?}

To answer this question, we develop a new framework referred to as {\em Learning Optimization Bias via Synthetic Task Generation (\textsc{OptBias})}. A key novelty of this work is the idea that offline optimizers can benefit from synthetic data generation, enabling training over a population of related functions rather than relying solely on the limited data available from the target task. This perspective introduces a new paradigm for offline optimization where augmented experience from related functions helps the optimizer navigate challenging OOD regions of the search space.
 Matching the surrogate’s gradient field to those of multiple related functions reinforces the recurring, shared structures while filtering out function-specific nuances. If the oracle function is indeed similar to this population, these shared structures are likely to align with its own, thus implicitly aligning the gradient field of the surrogate with that of the oracle. To substantiate this intuition, we need to address two technical challenges:  {\bf (1)}  How to generate auxiliary task functions which are similar to the oracle and synthesize strong training signals for gradient matching from them; and {\bf (2)} How to learn the optimization bias effectively from the synthetic training data?

{\bf Contributions.}~We address the above two challenges via the following technical contributions:

{\bf 1.~}We develop a robust procedure \textsc{Sim4Opt} to generate synthetic, closed-form functions which are similar to the oracle via fitting Gaussian process priors~\citep{Rasmussen06} on offline data and simulating input sequences with increasing output as learning feedback for gradient matching (Section~\ref{sec:sim-data}).  

{\bf 2.~}We develop a generalization of \textsc{Match-Opt}~\citep{hoang2025learning}, named \textsc{OptBias}, which leverages meta learning techniques to effectively incorporate such simulated gradient-matching feedback to synthesize a good approximation of the (unobserved) oracle's gradient field (Section~\ref{sec:meta-opt}).

{\bf 3.~}We conduct extensive experiments on real-world design optimization benchmarks to  demonstrate the effectiveness of \textsc{OptBias}. Our results show that it produces significantly better designs than baseline methods with small datasets. Our ablation experiments demonstrate that meta learning is a better approach than pretraining to capture optimization bias; and \textsc{OptBias} performs best when the simulated data are well-aligned with the optimization bias (Section~\ref{sec:exp}).

\section{BACKGROUND}

Let $\mathcal{X}$ denote the input design space and $g: \mathcal{X} \mapsto \Re$ be an unknown, expensive real-valued objective function that evaluates output $z$ = $g(\mathbf{x})$ for any input $\mathbf{x} \in \mathcal{X}$. The goal is to find an optimal design $\mathbf{x}_\ast = \argmax_{\mathbf{x}} \ g(\mathbf{x})$. Since the gradient of $g(\mathbf{x})$ is typically not accessible, previous approaches use derivative-free methods, such as random gradient estimation~\citep{WangAISTATS18}, or Bayesian optimization~\citep{SnoekICML15}, which requires online active sampling of experimental evaluations to either approximate the target function's derivative or learn its surrogate model. However, online data collection can be expensive or impractical. For example, lab-tests of candidate drugs are time-consuming and preferably done in high throughput setting, whereas test-driving an automatic controller for self-driving vehicles in real environments can be hazardous. Offline optimization has recently been proposed to avoid online data collection via fitting a parameterized surrogate model $g_\phi(\mathbf{x})$ for $g(\mathbf{x})$ on an existing dataset of past input designs and their evaluations, i.e., $\mathcal{D}=\{(\mathbf{x}_1, z_1), (\mathbf{x}_2, z_2), \cdots,(\mathbf{x}_n, z_T)\}$. This is achieved via solving the following:\vspace{-2mm}
\begin{eqnarray}
\phi &\triangleq& \underset{\phi'}{\argmin} \ \sum_{i=1}^n \ell\Big(g_{\phi'}\big(\mathbf{x}_i\big),\   z_i\Big) \ ,\label{eq:2}
\end{eqnarray}

\vspace{-2mm}
where $\phi$ denotes the parameters of surrogate model, e.g., $g_\phi(\mathbf{x}) = \phi^\top\mathbf{x}$, and $\ell$ denotes a supervised learning loss function, e.g., $\ell(z, z') = \|z-z'\|^2$. Once learned, $g_\phi$ is fixed and used as a surrogate to find the optimized designs via gradient search. This implicitly assumes that the target function is continuous. For design tasks on discrete input spaces, a continuous embedding of the data is first learned and offline optimization is performed on the embedding space. The focus of this paper is on real-world settings (e.g., drug design and materials discovery) where the size of offline experimental data is small ($T$ is a small number).

Since we employ Gaussian processes (GPs) to generate synthetic optimization tasks, 
we provide some background on GPs to make the paper self-contained.

\noindent {\bf Gaussian Processes.}~A Gaussian process (GP)~\citep{Rasmussen06} defines a prior distribution over a random function $g(\mathbf{x})$, characterized by a mean function $m(\mathbf{x})$ and a kernel function $k(\mathbf{x},\mathbf{x}')$.~Under this prior, let $\mathbf{z} = \{z_1, z_2, \ldots, z_n\}$ denote the set of noisy output $z_i = g(\mathbf{x}_i) + \epsilon$ with $\epsilon \sim \mathbb{N}(0, \sigma^2)$ at any finite set of inputs $\mathbf{X} = (\mathbf{x}_1,\mathbf{x}_2, \ldots, \mathbf{x}_n)$.~We have $\mathbf{x}$ follows a multivariate Gaussian distribution $\mathbf{z} \sim \mathbb{N}\big(\mathbf{m}, \, \mathbf{K}\big)$ where $\mathbf{m}=[m(\mathbf{x}_1),m(\mathbf{x}_2),\ldots m(\mathbf{x}_n)]$ denotes the mean vector, $\mathbf{K}$ denotes the covariance matrix with entries 
$\mathbf{K}^{ij} = k(\mathbf{x}_i, \mathbf{x}_j)$.~Given training data $D = (\boldsymbol{X}, \boldsymbol{z})$, the GP prediction $z_\ast$ and unseen test input $\mathbf{x}_\ast$ is also distributed by a Gaussian, $z_\ast \sim \mathbb{N}(\mu_D(\mathbf{x}_\ast), \sigma_D^2(\mathbf{x}_\ast))$ with:
\begin{eqnarray}
\hspace{-11mm}\mu_D(\mathbf{x}_\ast) &=& m(\mathbf{x}_\ast) + \mathbf{k}_\ast^\top\big(\mathbf{K} +\sigma^2\mathbf{I}\big)^{-1}\big(\mathbf{z}-\mathbf{m}\big) \\
\hspace{-11mm}\sigma^2_D(\mathbf{x}_\ast) &=& k(\mathbf{x}_*,\mathbf{x}_*) - \mathbf{k}_\ast^\top\big(\mathbf{K} + \sigma^2 \mathbf{I}\big)^{-1}\mathbf{k}_\ast \label{eq:gp}
\end{eqnarray}

where $\mathbf{k}_\ast = [k(\mathbf{x}_\ast,\mathbf{x}_1),\ldots,k(\mathbf{x}_\ast,\mathbf{x}_1)]$ denote the covariance vector between the test and training inputs.

\vspace{1ex}

\section{\textsc{OPTBIAS} FRAMEWORK}

In this section, we first provide the intuition and overview of the proposed \textsc{OptBias} framework. Next, we explain our \textsc{Sim4Opt} approach to generate simulated training data using a Gaussian process to address the small data challenge. Finally, we describe our meta learning approach to effectively learn optimization bias from simulated data using the gradient-matching procedure of ~\citep{hoang2025learning}.

\noindent {\bf Key Intuition and Overview of \textsc{OptBias}.}~Existing offline optimization methods often aim to learn surrogate models via value matching followed by a gradient based search for optimization. However, this is unnecessary for the purpose of optimization. For example, the maximizer of a function $g(\mathbf{x})$ is identical to that of $g(\mathbf{x}) \ +\  c$ for any constant $c$, which implies that preserving absolute function values is not required to recover the optimizer. Instead, it is more important that the surrogate preserves the gradient field of the oracle which leads to its optima. This motivates the use of gradient matching as a more appropriate optimization bias~\citep{hoang2025learning,tan2024offline} which focuses on learning the directional signals essential for optimization and avoids the unnecessary bias of matching absolute values.

\begin{figure}[t]
  \centering
  \includegraphics[width=\columnwidth]{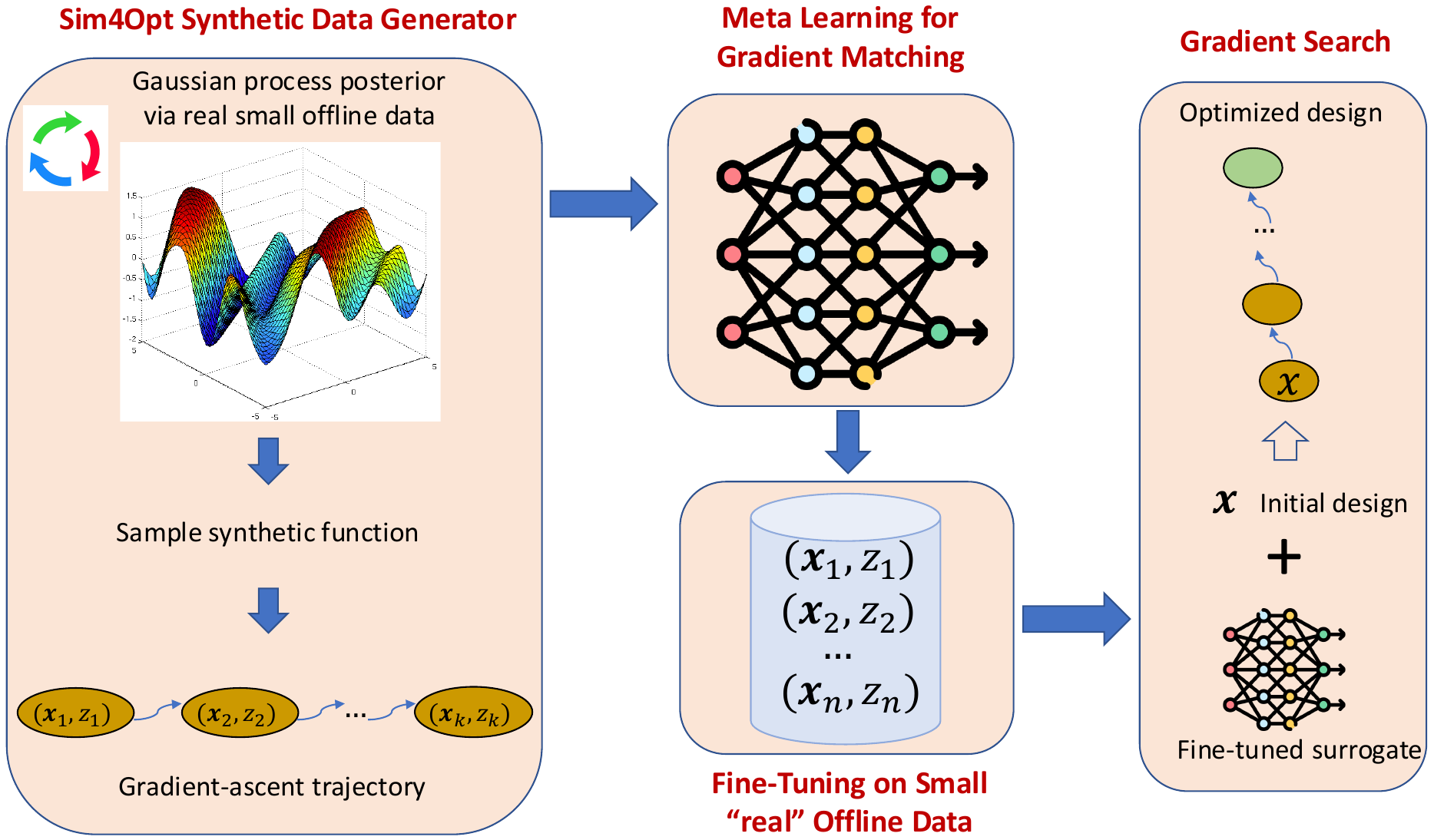}
\caption{Workflow overview of \textsc{OptBias}. First, we use a novel \textsc{Sim4Opt} procedure to generate synthetic functions which are similar to the oracle function. Then, we combine meta-learning with gradient matching (\textsc{MatchOpt}) to distill the common parts across gradient fields of the synthetic functions into a surrogate, thus implicitly aligning it with the (unobserved) gradient field of the oracle function.}
  \label{fig:workflow}
  \vspace{-2ex}
\end{figure}

However, learning to match gradient is bottlenecked by the lack of training data, which is often the case in real-world scenarios such as materials discovery and drug design.~To alleviate this bottleneck, we propose to enrich the offline data via synthetic data generation.~In particular, we aim to acquire additional learning feedback for gradient matching by generating and simulating gradient-ascent trajectories from a collection of synthetic functions similar to the oracle (Section~\ref{sec:sim-data}).~Our intuition is that by constructing synthetic functions similar to the oracle, their gradient fields will naturally overlap with the oracle’s.

Matching the surrogate’s gradient field to this collection of functions amounts to learning from the common structures that consistently recur across functions while down-weighting nuances that are specific to individual functions (Section~\ref{sec:meta-opt}).~In this way, the surrogate can recover shared patterns that are otherwise difficult to discern from limited offline data alone. Next, we fine-tune the learned meta-surrogate using the small (real) offline data for the target optimization task. Finally, we perform gradient search on the trained surrogate from high value inputs in the real offline data to find optimized designs.
{Notably, unlike prior work that avoids the oracle's out-of-distribution 
regime, \textsc{OptBias} introduces a new paradigm where synthetic data is used 
to navigate it more reliably.}
Figure~\ref{fig:workflow} illustrates key steps of \textsc{OptBias} and Algorithm~\ref{alg:optbias-min} gives the pseudo-code.

\vspace{-2ex}

\subsection{\textsc{Sim4Opt}: Simulated Data Generation}
\label{sec:sim-data}

One possible approach is to construct a variety of Gaussian process (GP) priors~\citep{Rasmussen06} over candidate functions and sample from them.~This idea was previously developed in ExPT~\citep{nguyen2023expt}, which samples GP priors by uniformly drawing from pre-specified parameter ranges. ~Each sampled GP prior defines a distribution over functions, from which synthetic functions can be drawn.~Data sampled from these functions are then used to supplement the limited offline dataset for surrogate training.~However, this approach suffers from the following key limitations:

{\bf 1.}~The GP priors used in ExPT are sampled from a pre-selected ranges rather than learned from offline data.~As such, there is a risk that some of which might be too distant from the target oracle function.~These can lead to synthetic functions whose gradient fields poorly align with the oracle’s.~This will in turn introduce spurious gradient structures that obscure the recurring gradient patterns across functions similar to the oracle that we aim to capture.

{\bf 2.}~ExPT generates synthetic data only by evaluating each sampled posterior mean function at unlabeled inputs drawn from the offline dataset.~Consequently, the resulting synthetic data might not correspond to any gradient-ascent trajectories that highlight the most informative regions of the gradient field.~Furthermore, the Design-Bench testbed~\citep{TrabuccoArXiv22} samples these inputs predominantly from low objective value regions of the oracle.~As those inputs lie in the oracle’s low objective value regime, they will likewise fall into the low objective value regions of the synthetic functions because the synthetic functions are constructed to resemble the oracle around the offline inputs.~As a result, the synthetic data remain confined to low objective value design spaces.~This limits the model’s ability to capture the common gradient structures across synthetic functions that point toward high objective value designs.  



To address these issues, we propose a new data generation method that can generate auxiliary functions which are more similar to the oracle function.~Furthermore, we also develop sampling techniques that draw highly informative gradient-ascent trajectories from these auxiliary functions.~The full data generation procedure is detailed in Algorithm \ref{alg:gen-data-multi}.

{\bf Addressing Limitation  1}. The key idea is to adapt the sampled GP prior in ExPT with the offline data via GP posterior computation to ensure that the resulting posterior has strong support over functions similar to the oracle.~The corresponding GP posterior's mean functions then have closed-form expressions which can be used to run $M$-step gradient descent and ascent from the offline input points.~This results in input trajectories that traverse between low objective value regions and high objective value regions, thus addressing the limitation. {This implicitly assumes that the oracle is a smooth function, or that at least its high-objective-value regime can be approximated well by a smooth function.} {Under this assumption, GP posterior mean functions capture similar geometric structures that connect low-value and high-value regions of the objective landscape.} 

{\bf Addressing Limitation  2}.~Such trajectories can be used to learn the common increasing objective value gradient structure across the sampled functions via customized meta learning approach (see Section~\ref{sec:meta-opt}). {Importantly, the GP is not used as a high-fidelity surrogate of the oracle. Instead, it serves only to generate a family of synthetic functions that behave similarly to the oracle within the offline data regime.} {Training the optimizer across this population of functions allows the meta-learner to capture optimization bias that generalizes to the target oracle despite the limited offline data.} {The central assumption is that if an optimizer performs well on average across a population of functions that are diverse but sufficiently similar to the oracle in the offline data regime, it will also perform well for the oracle itself.}~As these functions are similar to the oracle function, the common low objective value to high objective value gradient structure will likely align with the oracle's.~To better balance between exploration and exploitation in the simulation runs towards high-value regions, we can also replace the posterior mean with its upper-confidence bound (UCB) which was previously designed as an acquisition function in the context of online black-box optimization.

Indeed, we perform ablation experiments within \textsc{OptBias} by varying the synthetic data generator (our proposed \textsc{Sim4Opt} and the data generator in ExPT) to demonstrate the effectiveness of \textsc{Sim4Opt} procedure.


\begin{algorithm}[tb]
   \caption{\textsc{Sim4Opt}: Synthetic Data Generator}
   \label{alg:gen-data-multi}
\begin{algorithmic}[1]
   \STATE {\bfseries Input:} Offline dataset $D_o = \{\boldsymbol{X},\boldsymbol{z}\}$; evolution operator $\mathcal{T}$; number of evolution steps $M$; number of simulated functions $n$
   \STATE {\bfseries Output:} Mapping $\mathcal{S} = \{(i, D_s^{(i)})\}_{i=1}^n$, where each $D_s^{(i)} = \{(\boldsymbol{X}^{s,(i)}, \boldsymbol{z}^{s,(i)})\}$ is \emph{ordered order by} $\boldsymbol{z}^{s,(i)}$

   \STATE $\mathcal{S} \gets \emptyset$

   \FOR{$i=1$ {\bfseries to} $n$}
      \STATE $D_s^{(i)} \gets \emptyset$ \hfill // synthetic dataset for function $i$

      \STATE Sample kernel parameters for function $i$
      \STATE $\phi_s^{(i)} = (\ell_s^{(i)}, (\sigma_s^{2})^{(i)}) \sim U((\ell_0,\sigma_0^2)-\delta,(\ell_0,\sigma_0^2)+\delta)$ 
      \STATE Compute GP posterior mean $\bar{g}_{\phi_s^{(i)}}$ \textbf{via} Eq.~\eqref{eq:gp}

      \STATE Set $\boldsymbol{X}^{0,(i)}_+ \gets \boldsymbol{X}$, $\boldsymbol{X}^{0,(i)}_- \gets \boldsymbol{X}$

      \FOR{$m=1$ {\bfseries to} $M$}
         \STATE Explore low objective value regions:
         \STATE $\boldsymbol{X}^{m,(i)}_- \gets \mathcal{T}\!\left(-\,\bar{g}_{\phi_s^{(i)}},\, \boldsymbol{X}^{m-1,(i)}_- \right)$

         \STATE Explore high objective value regions:
         \STATE $\boldsymbol{X}^{m,(i)}_+ \gets \mathcal{T}\!\left(\ \bar{g}_{\phi_s^{(i)}},\, \boldsymbol{X}^{m-1,(i)}_+ \right)$

         \STATE Compute pseudo-labels
         \STATE $\boldsymbol{z}^{m,(i)}_- \gets \bar{g}_{\phi_s^{(i)}}(\boldsymbol{X}^{m,(i)}_-)$
         \STATE  $\boldsymbol{z}^{m,(i)}_+ \gets \bar{g}_{\phi_s^{(i)}}(\boldsymbol{X}^{m,(i)}_+)$

         \STATE \textbf{Update simulated data:}
         \STATE $D_s^{(i)} \gets D_s^{(i)} \cup \{\boldsymbol{X}^{m,(i)}_-, \boldsymbol{z}^{m,(i)}_-\}$
          \STATE $D_s^{(i)} \gets D_s^{(i)}\cup\ \{\boldsymbol{X}^{m,(i)}_+, \boldsymbol{z}^{m,(i)}_+\}$
      \ENDFOR

      \STATE \textbf{Monotone ordering (per function):} \STATE $D_s^{(i)} \gets \mathrm{SortByY}\!\left(D_s^{(i)}\right)$ \textit{(ascending in $\boldsymbol{z}^{s,(i)}$)}
      \STATE \textbf{Accumulate mapping:} $\mathcal{S} \gets \mathcal{S} \cup \{(i, D_s^{(i)})\}$
   \ENDFOR

   \STATE \textbf{Return} $\mathcal{S}$
\end{algorithmic}

\end{algorithm}
\vspace{-2ex}

\begin{algorithm}[tb]
  \caption{\textsc{OptBias} Algorithm}
  \label{alg:optbias-min}
\begin{algorithmic}[1]
  \STATE \textbf{Input:} Offline data $D_o$; simulated sets $\mathcal{S}=\{(i,D_s^{(i)})\}_{i=1}^n$ (Alg.~\ref{alg:gen-data-multi}); number of training epoch $E$; number of batches per epoch $n$,  stepsize $\alpha,\eta,\gamma$; 
  \STATE \textbf{Output: } Best designs $x^*$

  \STATE Initialize meta-learner hyperparameters $\phi$
  
  \STATE \textbf{Training meta-learner on simuated sets} 

  \FOR {epoch 1 to $E$}
      \STATE Sample a batch of $n$ functions $\{i,D^{(i)}_s\}_{i=1}^n\sim S$
      \FOR {$i$ in $1\ldots n$}
      \STATE Compute $\ell_i(\phi)$ via Eq.~\eqref{eq:matchopt-individual}
      \STATE Compute $\phi'_i=\phi-\alpha\nabla_\phi \ell_i(\phi)$
      \ENDFOR
      \STATE Update $\phi \gets \phi-\eta \nabla_\phi \sum_{i=1}^{n}\ell_i(\phi'_i)$
  \ENDFOR

  \STATE \textbf{Fine-tuning on small offline data $D_o$} 
  \STATE Compute $\ell(\phi)$ via Eq.~(\ref{eq:matchopt})
  \STATE Update $\phi \gets \phi - \alpha\nabla_\phi \ell(\phi)$
  
  \STATE \textbf{Gradient search on surrogate $\nabla_x g_\phi(x)$} 

  \FOR {t in {$1\ldots M$}}
    \STATE $\mathbf{x}_t \gets \mathbf{x}_{t-1} + \gamma \nabla_\mathbf{x} g_\phi(\mathbf{x}_{t-1})$
  \ENDFOR

\RETURN $\mathbf{x}^*$=$\mathbf{x}_M$
\end{algorithmic}
\end{algorithm}


\subsection{Meta Learning using Simulated Data}
\label{sec:meta-opt}
To remove the redundant value-matching bias, the gradient matching approach in~\citep{hoang2025learning} aims to learn a parameterized surrogate $g_\phi(\boldsymbol{x})$ that matches the oracle $g(\boldsymbol{x})$'s gradient field via $\phi^\ast = \argmin_\phi \ell(\phi)$ with the following gradient matching loss:\vspace{-2mm} 
\begin{eqnarray}
\hspace{-8mm}\ell(\phi)\hspace{-1mm}&=&\hspace{-1mm}\mathbb{E} \Bigg(\Delta z - \Delta\mathbf{x}^\top\hspace{-1mm} \int_0^1 \hspace{-1mm}\nabla g_\phi\Big(\mathbf{x} \ +\  t\Delta\mathbf{x}\Big)\mathrm{d}t\Bigg)^2 .\label{eq:matchopt}
\end{eqnarray}

\vspace{-4mm}
where $\Delta\mathbf{x} = \mathbf{x}' -\mathbf{x}$ and $\Delta z = g(\mathbf{x}') - g(\mathbf{x})$.~Eq.~\eqref{eq:matchopt} can be empirically estimated via sampling $(\mathbf{x}, \mathbf{x}')$ and their corresponding output values $(g(\mathbf{x}), g(\mathbf{x}'))$ from the offline dataset.~However, this is inaccurate with limited data.~To overcome this challenge, we will leverage the sampled gradient-ascent trajectories from our synthetic data generation procedure \textsc{Sim4Opt}, 
\begin{eqnarray}
\hspace{-6.5mm}\zeta^{(i)} \hspace{-2mm}&=&\hspace{-2mm} \Big(\left(\mathbf{x}^{(i)}_1, z^{(i)}_1\right), \left(\mathbf{x}^{(i)}_2, z^{(i)}_2\right), \ldots, \left(\mathbf{x}^{(i)}_\kappa, z^{(i)}_\kappa\right)\Big),    
\end{eqnarray}
with $\kappa = 2M + 1$ extracted from the synthetic functions obtained via Algorithm~\ref{alg:gen-data-multi}.~In particular, let $\ell_i(\phi)$ denotes the gradient matching loss with respect to $\zeta^{(i)}$ of the $i$-th GP posterior mean function,\vspace{-2mm} 
\begin{eqnarray}
\hspace{-2mm}\ell_i(\phi)\hspace{-2mm}&=&\hspace{-2mm}\mathbb{E}_r \Bigg(\Delta z_r - \Delta\mathbf{x}^\top_r\hspace{-1mm} \int_0^1 \hspace{-1mm}\nabla g_\phi\Big(\mathbf{x}_r +  t\Delta\mathbf{x}_r\Big)\mathrm{d}t\Bigg)^2\nonumber\label{eq:matchopt-individual}\\
\text{with}\ \Delta\mathbf{x}_r \hspace{-2mm}&=&\hspace{-2mm} \mathbf{x}^{(i)}_{r+1} - \mathbf{x}^{(i)}_r\ ,\Delta z_r \ =\  z^{(i)}_{r+1} - z^{(i)}_r \ .
\end{eqnarray}

\vspace{-2mm}
\noindent The expectation in Eq.~\eqref{eq:matchopt-individual} is over uniform choice of $r \in [\kappa - 1]$.~We can then learn a surrogate $g_\phi$ that matches with the common gradient-ascent structure across the sampled functions via minimizing the following meta-learning loss~\citep{Finn17,antoniou2018train},
\begin{eqnarray}
\phi^\ast &=& \argmin_\phi \mathbb{E}_i\Big[ \ell_i\Big(\phi - \alpha \nabla_\phi \ell_i(\phi)\Big) \Big ] \ ,\label{eq:meta-opt}   
\end{eqnarray}
where $\alpha > 0$ and $i \sim \mathbb{U}(1, n)$ assuming we have drawn $n$ synthetic functions.~Intuitively, this optimizes for a surrogate model whose gradient field can be fast-adapted towards any downstream target with limited data~\citep{Finn17}.~We will then fine-tune $\phi = \phi^\ast$ with the small (real) offline data for the target optimization task via Eq.~\eqref{eq:matchopt} to align with the oracle.~This intuitively emphasizes learning the common gradient fields across different functions along such gradient-ascent trajectories.~The solution of Eq.~\eqref{eq:matchopt-individual} can be intuitively viewed as a vantage point on the space of candidate gradient flows which can be fine-tuned quickly towards any downstream tasks.~Furthermore, as the gradient loss $\ell_i(\phi)$ is constructed along the sampled gradient-ascent trajectories $\zeta^{(i)}$, it naturally emphasizes fitting the gradient field along such low objective value to high objective value trajectories.


\vspace{-2ex}

\section{EXPERIMENTS AND RESULTS}
\label{sec:exp}

\begin{table*}[htbp]
\centering
\caption{Performance achieved by baselines under the limited data settings (using 1\% of the offline dataset) across various benchmark tasks.}
\label{tab:optbias_expt}
\resizebox{\linewidth}{!}{%
\begin{tabular}{|l||c|c|c|c|c|c|c|}
\toprule
& \multicolumn{7}{c|}{\textbf{Benchmarks}} \\
\cmidrule(lr){2-8}
\textbf{Method} & \textbf{Ant} & \textbf{D'Kitty} & \textbf{TFBind8}  & \textbf{RNA1} & \textbf{RNA2} & \textbf{RNA3} &\textbf{Mean Rank}\\
\midrule
$D_{best}$ \text{from 1$\%$ offline data} & 0.123 & 0.307 & 0.124  & 0.028 & 0.027 & 0.066 & None\\
\midrule
GA &  0.734 ± 0.054 &  0.831 ± 0.034  &  0.782 ± 0.117 &  0.475 ± 0.134 &0.497 ± 0.091 &  0.317 ± 0.072 & 5.17\\

MINs &  0.767 ± 0.084 &  0.895 ± 0.011 &  0.762 ± 0.135 & 0.076 ± 0.062 &  0.019 ± 0.029 &  0.132 ± 0.059 & 7.67 \\

COMs &  0.908 ± 0.042 &  0.705 ± 0.012 &  0.438 ± 0.000  & 0.279 ± 0.059 & 0.278 ± 0.120 & 0.293 ± 0.044 & 6.08\\

DEMO & 0.800 ± 0.150 & 0.845 ± 0.037 & 0.677 ± 0.168 & 0.166 ± 0.034 & 0.141 ± 0.045 & 0.295 ± 0.043& 6.50 \\

LTR & 0.865 ± 0.118 & 0.91 ± 0.026 & 0.45 ± 0.027 & 0.19 ± 0.189 & 0.215 ± 0.262 & 0.417 ± 0.176 & 5.17\\

Match-Opt & 0.859 ± 0.007 & 0.912 ± 0.013 & 0.438 ± 0.002 & 0.129 ± 0.000 & 0.119 ± 0.000 & 0.166 ± 0.000 & 6.92 \\

Batch BO (q-EI) &  0.482 ± 0.066 &  0.816 ± 0.030 &  0.928 ± 0.026 &  0.490 ± 0.114 &  0.523 ± 0.103 &  \textcolor{blue}{0.462 ± 0.097}& 4.00 \\

REINFORCE  &  0.327 ± 0.062 &  0.588 ± 0.162 &  0.872 ± 0.057 &  0.071 ± 0.054 &  0.062 ± 0.054 &  0.119 ± 0.064 & 8.83 \\

ExPT & 0.9276 ± 0.009 & \textcolor{blue} {0.955 ± 0.009} & 0.879 ± 0.080 & 0.249 ± 0.013 & 0.239 ± 0.009 & 0.341 ± 0.041 &  3.33\\
\midrule

\textbf{OptBias} (ours) & \textcolor{blue}{0.960 ± 0.017} & {0.947 ± 0.012} & \textcolor{blue}{0.945 ± 0.024} & \textcolor{blue}{0.503 ± 0.025} & \textcolor{blue}{0.540 ± 0.113} & {0.450 ± 0.134} & \textbf{1.33}\\
\bottomrule
\end{tabular}
}
\end{table*}

\begin{table*}[htbp]
\centering
\caption{Performance comparison between our proposed method \textsc{OptBias}, which uses meta-learning for surrogate training with synthetic data, and its variant, which replaces meta-learning with a pre-training procedure.}
\label{tab:optbias_surrogate}
\resizebox{\linewidth}{!}{%
\begin{tabular}{|l||c|c|c|c|c|c|}
\toprule
& \multicolumn{6}{c|}{\textbf{Benchmarks}} \\
\cmidrule(lr){2-7}
\textbf{Method} & \textbf{Ant} & \textbf{D'Kitty} & \textbf{TFBind8}  & \textbf{RNA1} & \textbf{RNA2} & \textbf{RNA3} \\
\midrule
\textbf{OptBias} (pre-training) & 0.884 ± 0.043 & 0.932 ± 0.007 & 0.438 ± 0.000 & 0.129 ± 0.000 & 0.119 ± 0.000 & 0.166 ± 0.000 \\
\textbf{OptBias} (meta-learning) & \textcolor{blue} {0.960 ± 0.017} & \textcolor{blue}{0.947 ± 0.012} & \textcolor{blue}{0.945 ± 0.024} & \textcolor{blue}{0.503 ± 0.025} & \textcolor{blue}{0.540 ± 0.113} & \textcolor{blue}{0.450 ± 0.134}\\
\bottomrule
\end{tabular}
}
\vspace{-3mm}
\end{table*}

This section presents the experiments and results to evaluate the performance of our proposed method \textsc{OptBias} in comparison to the baselines. 

\subsection{Benchmarks}

We conduct empirical evaluations on six benchmark tasks spanning robotics, genomics, and molecular biology. Each task is formulated as a black-box optimization problem, where the objective function is fixed and unknown to the optimizer. The setting follows the offline experimental design protocol: algorithms are provided only with a static dataset of design-response pairs for training and must recommend a batch of final designs for evaluation under the true objective. The black-box function is used solely at test time to assess the quality of these recommended designs.

The benchmark suite includes two continuous and one discrete tasks from the \texttt{Design-Bench} framework~\citep{trabucco2022design}, as well as three additional discrete RNA inverse folding tasks based on ViennaRNA~\citep{lorenz2011viennaRNA} and the Bootgen benchmark~\citep{kim2023bootstrapped}. A brief overview of each task is provided below:

\noindent \textbf{Ant Morphology (ANT)}~\citep{brockman2016openai}: A continuous task that involves optimizing the morphology of a quadruped robot to maximize forward locomotion speed. The input space is 60-dimensional and parameterizes the robot's body shape.
    
\noindent \textbf{D'Kitty Morphology (DKITTY)}~\citep{ahn2020robel}: A similar morphology optimization task for a different robot (D’Kitty), with a 56-dimensional continuous design space. The objective is to generate designs that improve locomotion or directional control.

\noindent \textbf{TF-Bind-8 (TF8)}~\citep{barrera2016survey}: This task focuses on optimizing DNA sequences of length 8 to maximize binding affinity to human transcription factors. The design space is discrete, consisting of strings over the nucleotide alphabet \{A, C, G, T\}.

\noindent \textbf{RNA-Binding Tasks (RNA-A, RNA-B, RNA-C)}~\citep{lorenz2011viennaRNA}: These inverse folding tasks aim to design RNA sequences (length 14) that fold into predefined target secondary structures. The goal is to maximize the predicted binding affinity by minimizing the free energy of the resulting structure. 

 These benchmarks collectively cover a wide range of structural properties (continuous vs. discrete design space, high vs. low dimensionality, providing a rigorous benchmark for evaluation. 
\subsection{Baselines}

\paragraph{Baselines.}
We compare against several recent baselines in offline optimization:
(1) \textbf{GA} which fits a surrogate using offline data and uses gradient ascend to find optimal input designs that maximize the surrogate output ~\citep{trabucco2022design}; 
(2) \textbf{MINs} which learns an \emph{inverse model} from scores to inputs (model inversion), so high-score conditions generate candidate designs directly, using only offline $(\mathbf{x},z)$ pairs~\citep{kumar2020model};
(3) \textbf{COMs} which learns a conservative (lower-bounding) surrogate to curb over-optimism on OOD inputs~\citep{trabucco2021conservative};
(4) \textbf{DEMO} which refines surrogate proposals with a diffusion prior to pull candidates toward the data distribution~\citep{yuan2024demo}; 
(5) \textbf{LTR (learning-to-rank)} which trains listwise/pairwise ranking losses to order candidates by quality rather than regress objective values~\citep{tan2024offline}; 
(6) \textbf{Match-Opt} which learns a differentiable surrogate whose \emph{gradient field} is trained to match improvement signals in the offline data~\citep{hoang2025learning}; 
(7) \textbf{Batch BO (q\!-EI)} which adopts batch Bayesian optimization using  batch expected improvement to recommend multiple designs per round;  
(8) \textbf{REINFORCE} which learns a variance-reduced score-function gradient estimator using importance sampling on offline data~\citep{williams1992simple}; 
and (9) \textbf{ExPT} which pre-trains a transformer on synthetic tasks and fine-tunes it with offline data of the target task~\citep{nguyen2023expt}.

\subsection{Evaluation Methodology} 
To evaluate the effectiveness of \textsc{OptBias} in scenarios with limited offline data, we consider a challenging few-shot setting constructed from each benchmark task. Specifically, we form the few-shot dataset $\mathcal{D}_{\mathrm{fs}}$ by selecting 1\% of lowest value data points (aka poorest) from the offline dataset to simulate a realistic scenario with limited experimental data similar to the setup in ExPT~\citep{nguyen2023expt}.~Following the approach in \citep{TrabuccoArXiv22}, each method generates $128$ optimized design candidates, which are then evaluated by the oracle function.~The candidates' performance are ranked and reported at 100th percentile.~All results are averaged over four independent runs. Specifically, each optimization method starts with $128$ initial candidates 
~The $128$ final candidate recommendations are then determined via updating the initial candidates following the corresponding optimization policy. In low-data regime, we score every unlabeled design from the offline data with our surrogate model, select the top 256 by predicted quality, and then randomly choose 128 of these as initial candidates.

\subsection{Results and Discussion}
This section evaluates \textbf{OptBias} across a diverse set of offline optimization tasks, including locomotion control (\texttt{ant}, \texttt{dkitty}), sequence design from the RNA inverse folding benchmark (\texttt{rna1}, \texttt{rna2}, \texttt{rna3}), and protein--DNA binding affinity prediction (\texttt{tfb-8}). Our experiments aim to answer the following questions:

\textbf{Q1.}~{\em How does \textsc{OptBias} compare to existing SOTA baselines in settings with limited data?}

\textbf{Q2.}~{\em What is the impact of using meta-learning to train a surrogate on synthetic data?}

\textbf{Q3.}~{\em What is the impact of generating data with \textsc{Sim4Opt} on optimization performance?}

{\textbf{Q4.}~{\em How sensitive is \textsc{OptBias} to the design choices of the synthetic task generation procedure in \textsc{Sim4Opt}?}

\textbf{Q5.}~{\em How well do synthetic designs generated by \textsc{Sim4Opt} capture high-value regions of the objective landscape?}}

We report the results and analyze the observations regarding each of the above questions below.


\begin{table*}[htbp]
\centering
\caption{Performance comparison between our proposed method \textsc{OptBias} and its variant which replaces \textsc{Sim4Opt} with \textsc{ExPT}'s data generation procedure.}
\label{tab:optbias_generation}
\resizebox{\linewidth}{!}{%
\begin{tabular}{|l||c|c|c|c|c|c|}
\toprule
& \multicolumn{6}{c|}{\textbf{Benchmarks}} \\
\cmidrule(lr){2-7}
\textbf{Method} & \textbf{Ant} & \textbf{D'Kitty} & \textbf{TFBind8} & \textbf{RNA1} & \textbf{RNA2} & \textbf{RNA3} \\
\midrule
OptBias w/ Random  & 0.701 ± 0.157 & {0.857 ± 0.058} & 0.882 ± 0.0548 & 0.493 ± 0.0458 & {0.532 ± 0.0549} & \textcolor{blue}{0.518 ± 0.077} \\
\textbf{OptBias} (ours) & \textcolor{blue} {0.960 ± 0.017} & \textcolor{blue}{0.947 ± 0.012} & \textcolor{blue}{0.945 ± 0.024} & \textcolor{blue}{0.503 ± 0.025} & \textcolor{blue}{0.540 ± 0.113} & {0.450 ± 0.134}\\
\bottomrule
\end{tabular}
}
\vspace{-3mm}
\end{table*}

\noindent {\bf Addressing Q1.}~Under limited-data settings (i.e., using 1\% of the original offline dataset), \textsc{OptBias} attains the best performance on 4/6 benchmark tasks.
~\textsc{OptBias} also achieves second-best performance on the remaining tasks: \textit{D’Kitty} (0.947 vs.~ExPT’s 0.955) and \textit{RNA3} (0.450 vs.~BO-qEI’s 0.462).~The performance gap in these tasks is however marginal.~Overall, these results demonstrate that incorporating synthetic data helps improve performance significantly over SOTA offline optimization methods in limited data settings.~The significant improvement of \textsc{OptBias} over \textsc{MatchOpt} also demonstrates that incorporating synthetic data is essential to improve gradient matching (to capture optimization bias).




\noindent {\bf Addressing Q2.}~To ablate the impact of using meta-learning in surrogate training, we compare the optimization performance achieved by \textsc{OptBias} and a variant that replaces meta-learning~\citep{finn2017maml} with a pre-training procedure that runs \textsc{MatchOpt} for gradient matching on data sampled from all synthetic tasks.~The key difference here is that meta-learning optimizes for a surrogate that can be fast-adapted to match the gradient field of any synthetic functions with limited data whereas the pre-training procedure attempts to match all gradient fields simultaneously.~Our hypothesis is that meta-learning is more effective than pre-training when using a low-complexity model, which is often preferable in offline optimization since the available data is often task-specific and moderate in size, thus cannot support pre-training a high-complexity model.~Intuitively, we can think of meta-learning as a mechanism that recognizes the most common parts across the gradient fields of different synthetic functions and filters out irrelevant parts, which makes fine-tuning easier on downstream target tasks.~Conversely, pre-training (regardless of model size) tends to overfit in settings with moderate and less diverse data.~This insight is supported by our empirical results in Table~\ref{tab:optbias_surrogate} which shows that \textsc{OptBias} consistently outperforms its variant that replaces meta-learning with pre-training.




\noindent {\bf Addressing Q3.}~To ablate the impact of generating data with \textsc{Sim4Opt}, we also compare \textsc{OptBias} with a variant which replaces \textsc{Sim4Opt} with the \textsc{ExPT}'s data generation produre for pre-training~\citep{nguyen2023expt}.~The key difference (as previously discussed in Section~\ref{sec:sim-data}) is that unlike \textsc{Sim4Opt}, \textsc{ExPT} generates synthetic data only within the offline input regime rather than along potential gradient-ascent trajectories that pass through the offline input regime.~Our hypothesis here is that as offline optimization is essentially a search from low objective value to high objective value regimes with the input space, observing explicit examples of trajectories with increasing objective function values that extend beyond the offline input regime is more informative than observing randomly within it.~This will in turn help improve fine-tuning in limited data setting (i.e., 1\% data), resulting in better optimization performance overall as reported in Table~\ref{tab:optbias_generation}.~It can be observed that \textsc{OptBias} (which uses \textsc{Sim4Opt} for data generation) is significantly better than its variant (that uses \textsc{ExPT}'s data generation procedure) on most tasks and on average.




{
\noindent {\bf Addressing Q4.}~To evaluate the robustness of Sim4Opt to design choices such as GP kernel selection, trajectory length, and the use of UCB vs.\ posterior mean, we conducted an ablation study across several commonly used configurations. The results summarized in Table~\ref{ablation:GP} show the final optimization performance from the OptBias approach over four seeds for each setting. Across ANT and DKitty, the variation in optimization performance across configurations is small relative to the overall gains provided by Sim4Opt. These results demonstrate that OptBias is stable fora wide range of reasonable hyperparameter choices for the synthetic task generation procedure.
}

{\color{red}
\begin{table*}[t]
\centering
\caption{Ablation study on GP design choices used in the Sim4Opt synthetic task generation procedure. Results show final optimization performance of OptBias averaged over four seeds.}
\begin{tabular}{lcc}
\toprule
Setting & ANT (mean $\pm$ std) & DKitty (mean $\pm$ std) \\
\midrule
Matern Kernel & 0.957 $\pm$ 0.004 & 0.898 $\pm$ 0.022 \\
UCB Acquisition & 0.937 $\pm$ 0.023 & 0.948 $\pm$ 0.010 \\
Trajectory Length = 70 & 0.942 $\pm$ 0.016 & 0.947 $\pm$ 0.009 \\
Lengthscale = 1.5 & 0.950 $\pm$ 0.008 & 0.871 $\pm$ 0.031 \\
Lengthscale = 2.0 & 0.916 $\pm$ 0.063 & 0.956 $\pm$ 0.007 \\
OptBias (main results) & 0.960 $\pm$ 0.010 & 0.947 $\pm$ 0.012 \\
\bottomrule
\end{tabular}
\label{ablation:GP}
\end{table*}
}

\begin{figure}[h]
\vspace{-3ex}
  \centering
\includegraphics[width=\columnwidth]{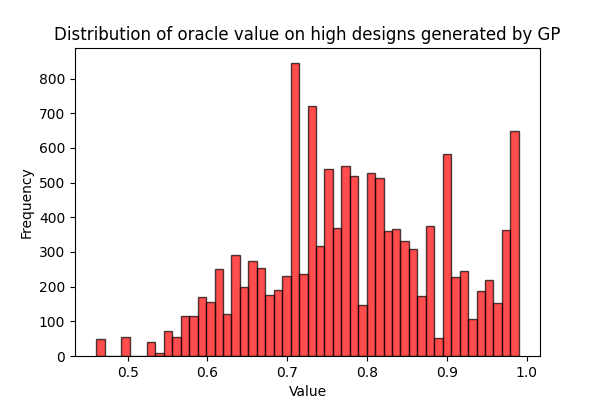}
  \caption{Empirical distribution of oracle-evaluated $y$-values (objective function values) over synthetic designs generated by multiple Gaussian Processes.}
  \label{fig:distribution}
\end{figure}

{
\noindent {\bf Addressing Q5.}
To evaluate the effectiveness of synthetic designs generated by multiple GPs, we first collect high pseudo-value candidates produced by these models. We then evaluate these candidates using the oracle function and analyze the resulting distribution of ground-truth objective values. As shown in Fig.~\ref{fig:distribution}, the synthetic designs exhibit a broad and informative distribution that sufficiently covers high-oracle-value regions. This coverage provides richer supervision signals, which are critical for addressing the out-of-distribution (OOD) challenge when performing gradient matching. In contrast, standard offline datasets are typically biased toward low-value regimes and lack sufficient high-quality samples. These observations strongly motivate our meta-training procedure on synthetic data, which enhances the model's ability to generalize beyond the support of the original dataset and effectively tackles the OOD problem.
}

\vspace{-2ex}

\section{RELATED WORK}

\vspace{-2ex}

\noindent {\bf Online Black-box Optimization.} Many classical approaches to experimental design for black-box optimization problems operate in an online manner: sequentially select a sequence of one or more experiments based on the outcomes of past experiments for optimization. Bayesian Optimization (BO) is the most prominent framework~\citep{garnett2023bayesian, snoek2012practical, swersky2013multi} for sample-efficient black-box function optimization. BO relies on two key elements. First, a surrogate model such as Gaussian Processes~\citep{srinivas2009gaussian,deshwal2021combining,deshwal2023bayesian}, Neural Processes~\citep{garnelo2018conditional, garnelo2018neural, gordon2019convolutional, kim2019attentive, nguyen2022transformer, singh2019sequential}, and Bayesian Neural Networks~\citep{goan2020bayesian} to make predictions with quantified uncertainty for unknown inputs. Second, a goal-driven acquisition function to 
to select one or more experiments. Some popular examples include expected improvement, upper confidence bound, and those based on information theory~\citep{garnett2023bayesian,ament2023unexpected,hernandez2014predictive,wang2017max,belakaria2019max,belakaria2020uncertainty}. 

However, in  real-world scenarios where the overhead and cost of setting up experiments are prohibitively expensive (e.g., wet lab experiments that require expensive materials and equipment), black-box optimization in the online setting is impractical. The closest BO approach to offline optimization is one-shot batch BO, e.g., q-EI~\citep{wang2020parallel}. Prior work in offline optimization and the experiments from this paper have demonstrated improvements over this BO baseline.

\noindent {\bf Offline Black-box Optimization.} This is an emerging and more practical paradigm which assumes access to an existing database of previously collected input-output pairs (aka offline dataset) and solves this problem in an offline manner. Existing methods can be broadly classified into two categories. First, {\em forward approaches} construct a surrogate model that approximates the true objective, and then optimize this proxy to identify promising candidates. The central challenge lies in ensuring reliable extrapolation for out-of-distribution (OOD) inputs. The most common strategy is to penalize large surrogate predictions on OOD inputs. There are several methods that implement this strategy using principles from adversarial robustness~\citep{trabucco2021conservative,yao2024gabo}, uncertainty quantification~\citep{trabucco2022design}, and semi-supervised learning~\citep{yuan2023ict,chen2023parallel}. Recent work has shown that learning relative rankings of inputs is a better objective to create effective surrogate models. To complement surrogate modeling, there are methods to create policies to guide gradient search towards high-performing inputs~\citep{chemingui2024pgs}. Second, {\em inverse approaches} primarily learn a conditional deep generative model that maps to potential designs, bypassing explicit objective modeling. Methods in this category differ in the choice of generative model: model inversion networks in MINs \citep{kumar2020model}, auto-regressive model in BONET \citep{BONET}, and diffusion model in DDOM, ROOT \citep{krishnamoorthy2023diffusion,dao2026root}. 

However, effectiveness of all these methods critically depends on large amounts of offline data and {\em small data} is the norm in many real-world applications which is the focus of this paper. A very recent work referred to as ExPT~\citep{nguyen2023expt} addresses the small data challenge using pre-training on synthetic data generated from a GP with random hyper-parameters but does not capture the optimization bias important for strong performance. Our \textsc{OptBias} framework achieves this using meta learning with data from an improved synthetic data generation procedure.

\vspace{-2ex}

\section{CONCLUSION}

This paper studied the under-studied problem of black-box optimization from small offline datasets which is common in many scientific applications. It addressed the challenge of small data setting using a novel synthetic data generation procedure and meta learning to capture optimization bias within a meta surrogate which is then fine-tuned on the small real data from the target task. Our experiments and ablations demonstrate the effectiveness of both synthetic data generator and meta learning approach, resulting in improvements over state-of-the-art baselines. {Exploring alternative meta-learning approaches within \textsc{OptBias} is an interesting direction for future work.}

\bibliographystyle{plainnat}
\bibliography{aistat,Nghia-bib,Jana-bib}


\section*{Checklist}

\begin{enumerate}

  \item For all models and algorithms presented, check if you include:
  \begin{enumerate}
    \item A clear description of the mathematical setting, assumptions, algorithm, and/or model. [Yes]
    \item An analysis of the properties and complexity (time, space, sample size) of any algorithm. [Yes]
    \item (Optional) Anonymized source code, with specification of all dependencies, including external libraries. [Yes]
  \end{enumerate}

  \item For any theoretical claim, check if you include:
  \begin{enumerate}
    \item Statements of the full set of assumptions of all theoretical results. [Not Applicable]
    \item Complete proofs of all theoretical results. [Not Applicable]
    \item Clear explanations of any assumptions. [Not Applicable]     
  \end{enumerate}

  \item For all figures and tables that present empirical results, check if you include:
  \begin{enumerate}
    \item The code, data, and instructions needed to reproduce the main experimental results (either in the supplemental material or as a URL). [Yes]
    \item All the training details (e.g., data splits, hyperparameters, how they were chosen). [Yes]
    \item A clear definition of the specific measure or statistics and error bars (e.g., with respect to the random seed after running experiments multiple times). [Yes]
    \item A description of the computing infrastructure used. (e.g., type of GPUs, internal cluster, or cloud provider). [Yes]
  \end{enumerate}

  \item If you are using existing assets (e.g., code, data, models) or curating/releasing new assets, check if you include:
  \begin{enumerate}
    \item Citations of the creator If your work uses existing assets. [Yes]
    \item The license information of the assets, if applicable. [Not Applicable]
    \item New assets either in the supplemental material or as a URL, if applicable. [Yes]
    \item Information about consent from data providers/curators. [Not Applicable]
    \item Discussion of sensible content if applicable, e.g., personally identifiable information or offensive content. [Not Applicable]
  \end{enumerate}

  \item If you used crowdsourcing or conducted research with human subjects, check if you include:
  \begin{enumerate}
    \item The full text of instructions given to participants and screenshots. [Not Applicable]
    \item Descriptions of potential participant risks, with links to Institutional Review Board (IRB) approvals if applicable. [Not Applicable]
    \item The estimated hourly wage paid to participants and the total amount spent on participant compensation. [Not Applicable]
  \end{enumerate}

\end{enumerate}

\appendix

\onecolumn
\section*{APPENDIX}
\subsection*{A. ADDITIONAL PERFORMANCE COMPARISON WITH OTHER BASELINES}
In addition to the 9 baseline methods in the main text, we further compare our method with 6 recent offline optimization techniques:
(1) \textbf{DDOM} which learns a conditional diffusion generator \(p_\theta(x\mid y)\) that enables guided sampling to generate high-score design candidates~\citep{krishnamoorthy2023diffusion};
(2) \textbf{Parallel-mentoring (Tri-mentoring)} which learns three proxies to mentor each other via voting-based pairwise supervision and soft labels to strengthen the ensemble and curb OOD errors \citep{chen2023parallel};
(3) \textbf{BDI} enforces a forward and backward mapping between static datasets and high-scoring designs using neural tangent kernel theory \citep{chen2022bidirectional}; (4) \textbf{RGD} combines explicit proxy guidance with proxy-free diffusion modeling and iteratively refines the proxy via diffusion feedback \citep{chen2024robust}, { (5) \textbf{GTG} generates design trajectories by guiding diffusion models along high-value paths for offline optimization \citep{yun2024guided}; and (6) \textbf{PGS} reinterprets offline optimization as a reinforcement learning task, optimizing for an effective policy using sampled trajectories from offline data \citep{chemingui2024pgs}.} Other methods, such as IGNITE and BOSS~\citep{dao2024incorporating,dao2025boosting}, are boosting frameworks that can be incorporated into arbitrary surrogate models. Therefore, we do not include these boosted variants as direct comparison baselines.

\begin{table*}[htbp]
\centering
\caption{Performance achieved by baselines under the limited data settings (using 1\% of the offline dataset) across various benchmark tasks.}
\label{tab:optbias_expt}
\resizebox{0.8\linewidth}{!}{%
\begin{tabular}{|l||c|c|c|c|}
\toprule
& \multicolumn{4}{c|}{\textbf{Benchmarks}} \\
\cmidrule(lr){2-5}
\textbf{Method} & \textbf{Ant} & \textbf{D'Kitty} & \textbf{TFBind8} & \textbf{Mean Rank} \\
\midrule
$D_{best}$ \text{from 1$\%$ offline data} & 0.123 & 0.307 & 0.124 & None \\

\midrule
DDOM & 0.324  ± 0.024 &  0.746 ± 0.012  & 0.771  ± 0.121 & 5.33 \\

Tri-mentoring & 0.831  ± 0.015 &  0.928 ± 0.008 & 0.645  ± 0.030 & 4.0 \\

BDI &  0.298 ± 0.000  &  0.728 ± 0.000  &  0.447 ± 0.000 & 5.67 \\

RGD & 0.870 ± 0.022 & 0.888 ± 0.015 & 0.890 ± 0.057 & 3.17 \\

GTG &  0.825 ± 0.050 & 0.913 ± 0.020 & 0.890 ± 0.054 & 3.17\\

PGS & 0.572 ± 0.021 & 0.882 ± 0.013 & 0.536 ± 0.051 & 4.83 \\

\midrule
\textbf{OptBias (ours)} & \textcolor{blue}{0.960 ± 0.017} & \textcolor{blue}{0.947 ± 0.012} & \textcolor{blue}{0.945 ± 0.024} & \textcolor{blue}{1.00} \\
\bottomrule
\end{tabular}
}
\end{table*}

\subsection*{B. HYPERPARAMETERS}
\noindent We present our hyperparameter setting in this section. In our synthetic data generator (Alg.~\ref{alg:gen-data-multi}), we generate $n=128$ simulated functions, and use the evolution operator $\mathcal{T}$ as gradient ascent with $100$ gradient steps. The learning rate of this gradient process is $0.05$ for continuous tasks and $0.005$ for discrete tasks. The evolution operator $\mathcal{T}$ could be set as any other optimization method, such as a genetic program or local search.~We currently use gradient search due to its simplicity.~After the generation phase, we trained our meta-learner for $E$ epochs, with $E=50$ for continuous tasks and $E=100$ for discrete tasks, using a learning rate of $\eta=0.001$. After training, we fine-tuned the model for 20 epochs and then leveraged the fine-tuned model for gradient search with a step size of $\gamma=0.001$ for continuous tasks, $\gamma=0.1$ for discrete tasks, and 300 steps to obtain our optimal designs. We provide a comprehensive report of the hyperparameters used in our implementation for both continuous and discrete tasks.

\begin{table}[htbp]
\centering
\small
\begin{minipage}[t]{0.48\textwidth}
\centering
\textbf{Continuous (e.g., AntMorphology)}\\[2pt]
\begin{tabular}{l l}
\toprule
\textbf{Setting} & \textbf{Value} \\
\midrule
Num context / target & 16 / 64 \\
Batch size & 128 \\
Epochs ($E$) & 50 \\
Inner-loop learning rate & 0.1 \\
Meta learning rate & 0.001 \\
Norm layer & BatchNorm \\
\midrule
\multicolumn{2}{l}{\textit{Meta-training function generator (GP)}} \\
Num functions ($n$) & 128 \\
Kernel & RBF \\
Init lengthscale / variance & 1.0 / 1.0 \\
Noise & 0.01 \\
Top-$k$ fraction (poorest data) & 0.01 \\
Inner gradient steps & 100 \\
Learning rate (GP sim) & 0.05 \\
\midrule
\multicolumn{2}{l}{\textit{Finetuning / gradient search}} \\
Finetuning epochs & 20 \\
Gradient search iterations & 300 \\
Stepsize ($\gamma$) & 0.001 \\
\bottomrule
\end{tabular}
\end{minipage}\hfill
\begin{minipage}[t]{0.48\textwidth}
\centering
\textbf{Discrete (e.g., TFBind-8)}\\[2pt]
\begin{tabular}{l l}
\toprule
\textbf{Setting} & \textbf{Value} \\
\midrule
Num context / target & 16 / 64 \\
Batch size & 128 \\
Epochs ($E$) & 100 \\
Inner-loop learning rate & 0.1 \\
Meta learning rate & 0.001 \\
Norm layer & BatchNorm \\
\midrule
\multicolumn{2}{l}{\textit{Meta-training function generator (GP)}} \\
Num functions ($n$) & 128 \\
Kernel & RBF \\
Init lengthscale / variance & 6.25 / 6.25 \\
Noise & 0.01 \\
Top-$k$ fraction (poorest data) & 0.01 \\
Inner gradient steps & 100 \\
Learning rate (GP sim) & 0.005 \\
\midrule
\multicolumn{2}{l}{\textit{Finetuning / gradient search}} \\
Finetuning epochs & 25 \\
Gradient search iterations & 300 \\
Stepsize ($\gamma$) & 0.1 \\
\bottomrule
\end{tabular}
\end{minipage}
\caption{Hyperparameters for continuous (left) and discrete (right) tasks.}
\end{table}

\subsection*{C. ABLATION STUDIES}

\noindent We vary the number of GP-sampled functions used during meta-training, $K \in {8,16,32,64,128}$. Each function defines a different optimization task, so the meta-learner must focus on patterns shared across tasks rather than on task-specific details. Sets are nested (smaller $K$ is a subset of larger $K$). \emph{Mean Rank} averages over tasks with values for all methods. Performance stabilizes for $K \ge 64$, with only small gains at $K=128$ (see Table 6 in Appendix), showing that a modest number of synthetic functions is enough.

\begin{table*}[htbp]
\centering
\small
\resizebox{\linewidth}{!}{%
\begin{tabular}{lccccccc}
\toprule
\textbf{Method ($K$)} & \textbf{Ant} & \textbf{D'Kitty} & \textbf{TFBind8} & \textbf{RNA1} & \textbf{RNA2} & \textbf{RNA3} & \textbf{Mean Rank} \\
\midrule
OptBias (128) & \textbf{0.960 ± 0.017 } \,(\textbf{1}) & \textbf{0.947 ±  0.012} \,(\textbf{1}) & \textbf{0.945 ± 0.024 } \,(\textbf{1}) & 0.503 ± 0.025 \,(2) & \textbf{0.540 ± 0113 } \,(\textbf{1}) & 0.450 ± 0.134\,(4) & \textbf{1.67} \\
OptBias (64)  & 0.912 ± 0.082 \,(3) & 0.938 ± 0.029 \,(2) & 0.916 ± 0.087\,(2) & \textbf{0.541 ± 0.149 } \,(\textbf{1}) & 0.531 ± 0.115 \,(2) & \textbf{0.550 ± 0.035} \,(\textbf{1}) & 1.83 \\
OptBias (16)  & 0.926 ± 0.060 \,(2) & 0.906 ± 0.050 \,(3) & 0.855 ± 0.080 \,(4) & 0.437 ± 0.137 \,(3) & 0.453 ± 0.136 \,(3) & 0.539 ± 0.044 \,(2) & 2.83 \\
OptBias (8)   & 0.823 ± 0.130 \,(4) & 0.890 ± 0.081 \,(4) & 0.871 ± 0.060 \,(3) & 0.368 ± 0.214 \,(4) & 0.377 ± 0.183 \,(4) & 0.471 ± 0.088 \,(3) & 3.67  \\
\bottomrule
\end{tabular}}
\caption{Performance of \textbf{OptBias} under different $K$ across Design-Bench tasks. Higher is better. Bold marks the best value (and its rank) per task, and the best mean rank. }

\label{tab:optbias_k}
\end{table*}

Overall, increasing the number of meta-training functions ($K$) consistently improves performance across tasks, indicating more diverse training functions enhance generalization.

\subsection*{D.  MODEL ARCHITECTURE}
We use a lightweight meta-learning regression network: a small feed-forward stack \textbf{with hidden layers [512, 128, 32]} and normalization + LeakyReLU activations, followed by a linear head to a single scalar; it supports externally supplied (inner-loop) parameters and step-specific normalization statistics for fast adaptation across tasks.

\subsection*{E. SCALABILITY OF GAUSSIAN PROCESSES WITHIN \textsc{SIM4OPT}}

{Gaussian Processes are known to scale cubically with the number of training points. However, the focus of this work is on data-scarce offline optimization settings where the available dataset is typically small. In such regimes the computational cost remains manageable even in moderately high-dimensional spaces. For example, the datasets used in our experiments already have relatively high input dimensions: the ANT task uses 60-dimensional inputs ($d=60$) and the DKitty task uses 56-dimensional inputs ($d=56$). The training complexity of GPs is $O(n^3 + dn^2)$, where $n$ is the dataset size and $d$ is the input dimension, which scales linearly with respect to $d$.
For larger datasets, scalable GP approximations such as sparse Gaussian Processes can be used to further reduce computational cost. These methods approximate the full covariance structure using a set of inducing points, allowing the training complexity to scale approximately linearly with the number of data points while maintaining strong predictive performance \citep{hensman2013gaussian}.}

\subsection*{F. GPU COMPUTE FOR EXPERIMENTS}
All experiments were run on a single \textbf{NVIDIA A40 (46\,GB)} with \textbf{CUDA 12.9} and driver \textbf{575.51.03}.

\subsection*{G. CODE AVAILABILITY}
The official implementation of this work is publicly available at:
\url{https://github.com/azzafadhel/OptBias.git}

\onecolumn

\end{document}